\documentclass[acmtog]{acmart}
\usepackage{booktabs}
\usepackage{graphicx}
\usepackage{caption}
\usepackage{url}

\setcopyright{none}
\renewcommand\footnotetextcopyrightpermission[1]{}
\makeatletter
\let\@authorsaddresses\@empty
\makeatother
\title{Text-Driven Artistic Staging: 3D Posing, Lighting, and Camera References from Paintings}

\definecolor{duckyellow}{RGB}{218,165,32}

\subtitle{%
  \textcolor{duckyellow}{%
    \textbf{Live Demo:} \url{https://artstage.darkmesh.art/}%
  }%
}

\author{Yunge Wen}
\affiliation{%
  \institution{Massachusetts Institute of Technology, New York University}
\city{Cambridge}
  \state{Massachusetts}
  \country{USA}
}
\email{yungew@mit.edu}

\begin{document}

\begin{abstract}
Artists coordinate human pose, illumination, and camera placement to convey narrative and emotion, but existing generative methods typically model these elements independently. We introduce text-to-editable 3D staging, a task that jointly generates human poses, a dominant light, and a camera configuration from an affective description. We construct 11,911 text--staging pairs from 2,328 figurative paintings by reconstructing SMPL bodies, estimating low-frequency illumination, recovering camera parameters, and pairing each scene with ArtEmis descriptions. We train a flow-matching transformer that supports variable numbers of figures and produces multiple staging alternatives for each prompt. On held-out descriptions, the model achieves 32.2\% retrieval R@1, compared with 16.6\% for CLIP-based nearest-neighbor retrieval, while approximately preserving corpus-level diversity. These results demonstrate the feasibility of generating editable, emotionally conditioned 3D staging references from text.
\end{abstract}

\begin{teaserfigure}
  \includegraphics[width=\textwidth]{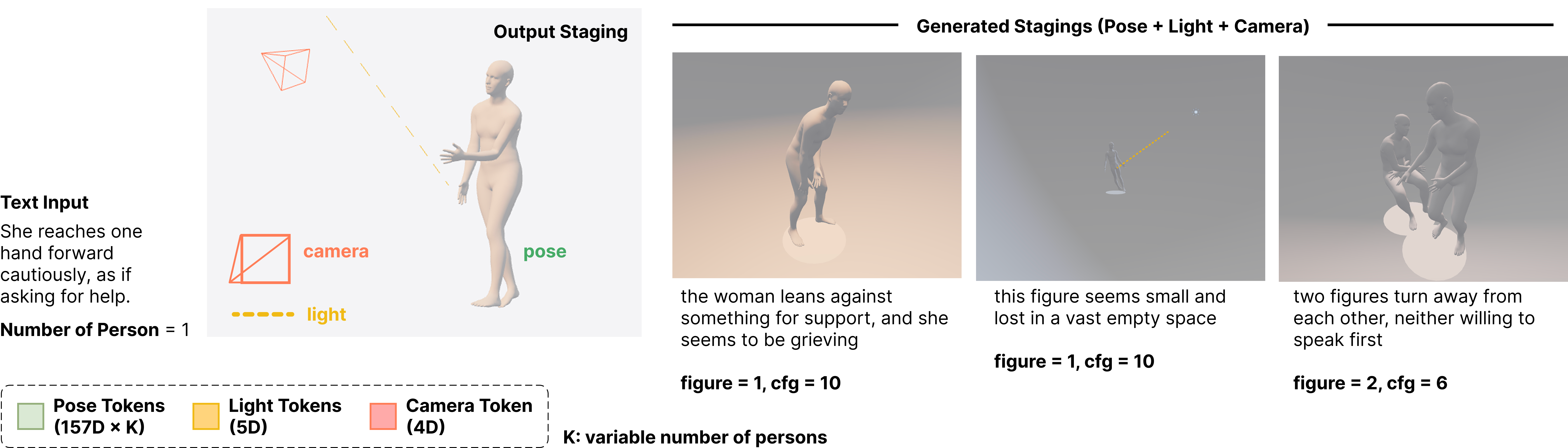}
  \caption{A staging is a pose for every figure, a light and a camera,
  generated together from one sentence and editable afterwards. Left: the three
  components of a single sample, from a description that names the body
  explicitly. Right: samples for three affective descriptions of the kind
  ArtEmis annotators actually write, which name a feeling and leave the body
  and the framing to the model; one has two figures.}
  \label{fig:teaser}
\end{teaserfigure}

\maketitle

% \captionsetup{skip=2pt}
\setlength{\textfloatsep}{2pt plus 1pt minus 1pt}
\setlength{\floatsep}{2pt plus 1pt minus 1pt}

\section{Introduction}
Composition and layout are foundational to painting, storyboarding, theatre staging and photography alike. Staging an emotionally charged figure, a human artist intuitively settles three things at once: how the body is held, where the light falls, and where the viewer stands. A low viewpoint reads as looming only if the figure is upright; a rim light isolates only if the pose turns away. Choosing any one commits the other two.

Existing tools address one component at a time. Pose libraries and 3D mannequins model the body and leave lighting and framing to the artist. Generative methods are partitioned the same way: text-to-pose and text-to-motion work outputs the body alone~\cite{feng2024chatpose,tevet2023mdm}, camera trajectory methods~\cite{courant2024et,jiang2024ccd} take character motion as a conditioning \emph{input} and do not model light, and relighting for images and video~\cite{kocsis2024lightit} produces images rather than a 3D scene. To our knowledge, existing methods do not jointly generate an editable staging from text describing the intended emotion or narrative.

We recover the joint solution from figurative painting, where pose, lighting and camera were resolved together to make a feeling legible. We contribute (i) a corpus of 11{,}911 (emotional text, 3D staging) pairs reconstructed from paintings and paired with ArtEmis~\cite{achlioptas2021artemis} annotator responses; (ii) a flow-matching transformer that generates pose, lighting and camera jointly from one sentence, giving several distinct alternatives per query; and (iii) an evaluation showing that guidance trades text alignment against variation.

\begin{figure}
    \centering
    \includegraphics[width=0.75\linewidth]{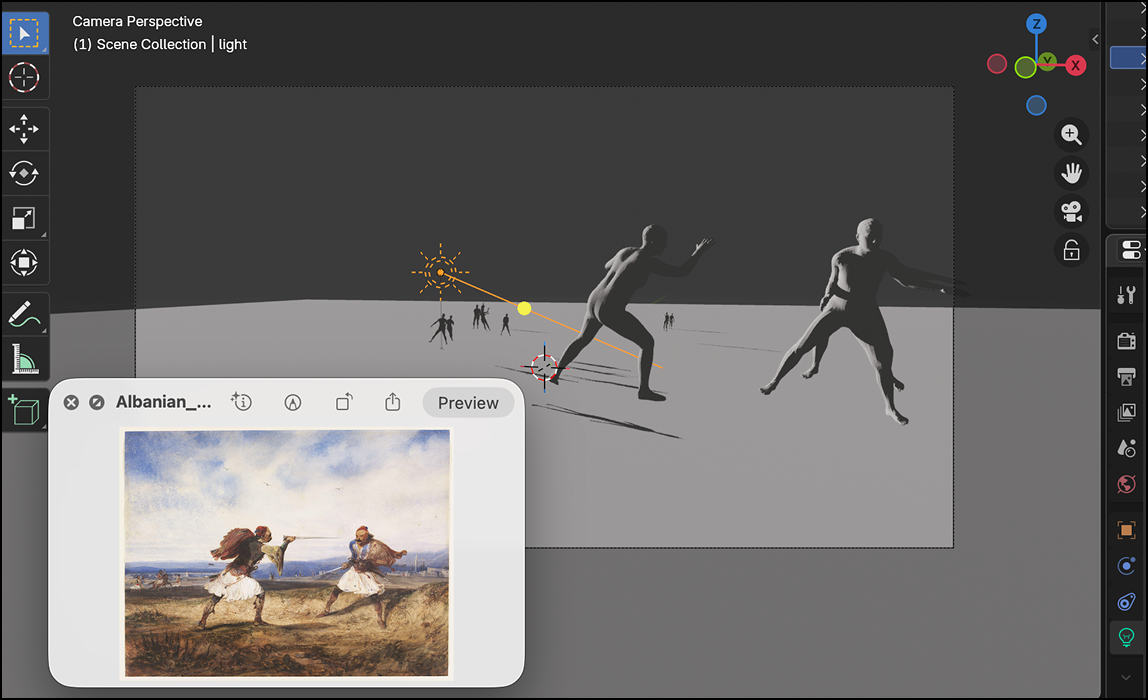}
    \caption{A recovered staging beside the painting it came from. Every
    figure is an SMPL body, the light is a direction and a strength fitted to
    the heads, and the camera is the reconstruction viewpoint; the result is a
    scene an artist can open and edit.}
    \label{fig:dataset}
\end{figure}

\section{Dataset}
From the intersection of ArtEmis~\cite{achlioptas2021artemis} and WikiArt, we retain paintings that serve as professionally composed staging examples associated with an emotion agreed upon by most annotators and whose style supports reliable human reconstruction. Mask R-CNN then discards paintings with no person, figures that are too small, or more than four figures, leaving 6{,}826 candidates.

\textbf{Pose.} Each detected box goes to HMR2~\cite{goel2023humans4d} for an SMPL~\cite{loper2015smpl} body; the projected silhouette against the detector's mask scores the fit.

\textbf{Lighting.} Visible SMPL vertices within $0.11$\,m of the head centroid are sampled from the image, and least squares over the resulting (normal, luminance) pairs fits the first-order spherical harmonic $\text{luminance}=c_0+\mathbf{c}_1\!\cdot\!\mathbf{n}$. The direction and magnitude of $\mathbf{c}_1$ estimate the dominant low-frequency illumination and its directional variation. Per-channel fits estimate illumination chromaticity, from which we derive a correlated colour temperature.

\textbf{Camera.} Position follows from the reconstruction, with the group's lowest vertex on the ground. Field of view is fixed at $45^{\circ}$. For multiple figures, the 4DHumans crop-to-full conversion maps each reconstruction into a shared full-image camera coordinate system.

Filtering on these scores (Fig.~\ref{fig:dataset}) keeps 2{,}328 paintings, 34\% of candidates, silhouette IoU and lighting fit accounting for almost all of the loss. Up to six ArtEmis sentences per work, each encoded by CLIP~\cite{radford2021clip} into a token sequence, give $N=11{,}911$ description--staging pairs.

\begin{figure}
    \centering
    \includegraphics[width=1\linewidth]{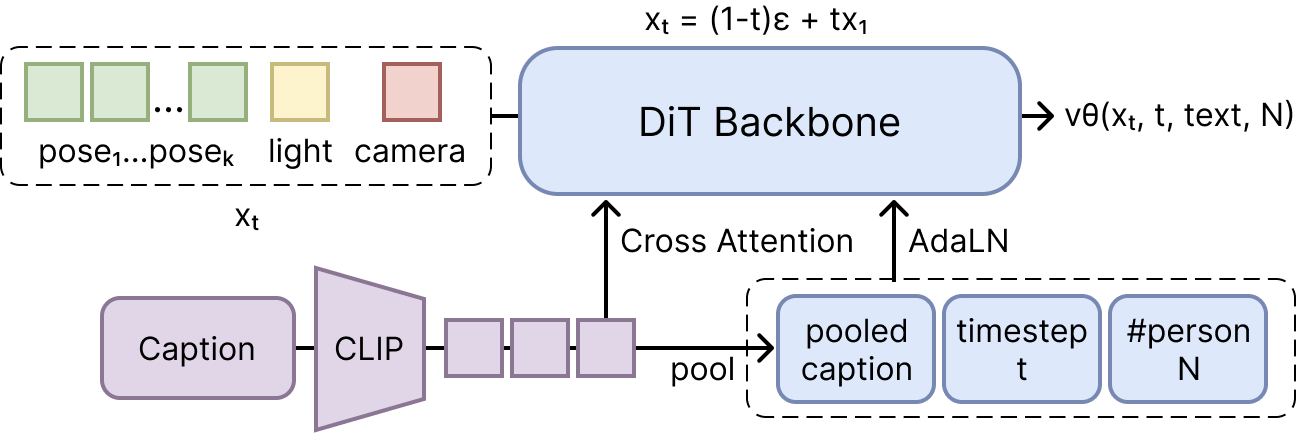}
    \caption{Model Architecture}
    \label{fig:arch}
\end{figure}

% \begin{figure}
%     \centering
%     \includegraphics[width=1\linewidth]{figures/guidance.png}
%     \caption{One description, one noise sample, three guidance weights. As $w$
%     rises the arm comes up and the camera swings onto the axis of the light,
%     the description taking over from the noise; $w\!=\!6$ is the weight we report
%     on, $w\!=\!11$ the upper end of the sweep in Fig.~\ref{fig:sweep}.}
%     \label{fig:guidance}
% \end{figure}

\section{Method}
\textbf{Representation.} A staging is a sequence of typed vectors $[\,p_1 \dots p_K,\ \ell,\ c\,]$ for $K\!\le\!4$ figures: $p_i\in\mathbb{R}^{157}$ is a body ($24\times6$ joint rotations, $10$ SMPL shape coefficients, $3$ root position), $\ell\in\mathbb{R}^{5}$ the light (a unit direction as $xyz$, directional strength, colour temperature), $c\in\mathbb{R}^{4}$ the camera (elevation, azimuth as sine and cosine, log distance). Typed tokens let scenes with different figure counts share one model without padding, and let the figures attend to each other. The two quantities that wrap, camera azimuth and light direction, are stored as sine and cosine or as a unit vector; elevation is bounded and kept as an angle, and distance is stored as its logarithm. Sagittal reflection augments half of each epoch, mirroring the light and camera azimuth alongside the body.

\textbf{Model.} A pre-norm transformer of four layers at width 256 (Fig.~\ref{fig:arch}). The CLIP-encoded description enters through cross-attention as a token sequence rather than pooled, so that the camera and light tokens can attend to the words describing them.

\textbf{Objective.} Flow matching~\cite{liu2022rectified}, as in recent camera-control work~\cite{bahmani2025ac3d}: with $x_t=(1-t)\,\epsilon+t\,x_1$ the network regresses $x_1-\epsilon$, sampled in 30 Euler steps. Descriptions are dropped 10\% of the time, making the guidance weight $w$ available at sampling time. Training uses dropout 0.2, EMA and early stopping on a scene-level split.

\section{Evaluation}
Following text-conditioned camera generation~\cite{courant2024et,jiang2024ccd}, we train and freeze a contrastive language--staging encoder. We measure alignment by R@1 over 32 held-out descriptions (3.125\% chance), diversity across descriptions, and multimodality (MM) across repeated samples. As a direct test against corpus reuse, retrieval selects the staging whose training caption is nearest to the query in frozen CLIP space and has zero MM by construction.

\begin{figure}[t]
  \centering
  \includegraphics[width=\linewidth]{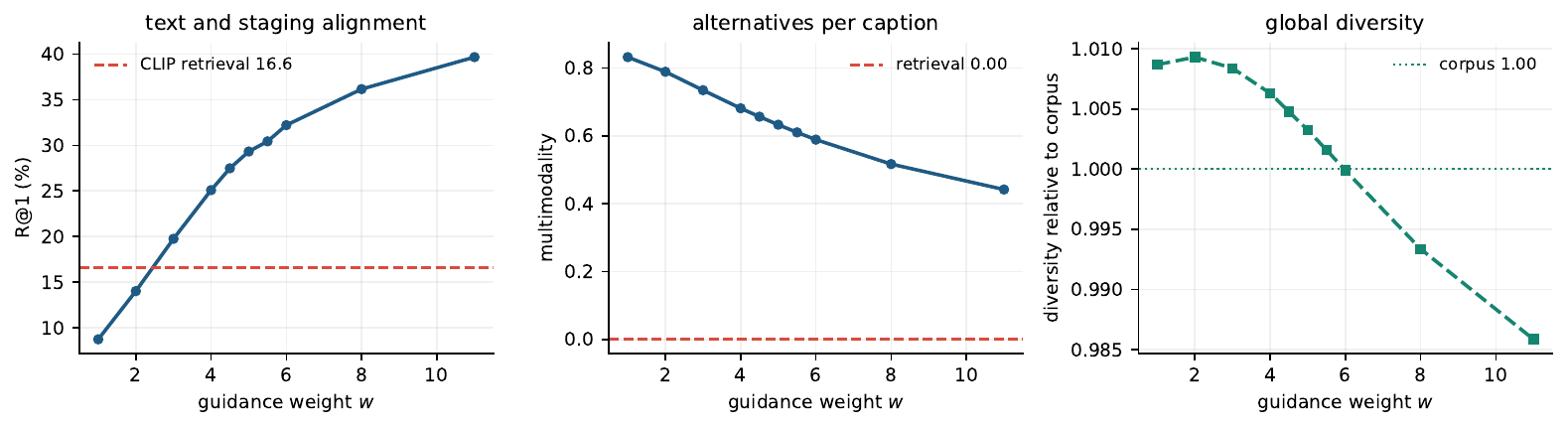}
  \caption{Effect of guidance weight. Text alignment increases as $w$ rises,
  while alternatives per description decrease. The model surpasses CLIP retrieval
  at $w\!\approx\!2.5$ and approximately matches corpus-level diversity at
  $w\!=\!6$. Retrieval has zero multimodality because it returns one staging
  per description.}
  \label{fig:sweep}
\end{figure}

We ablate classifier-free guidance in Fig.~\ref{fig:sweep}. Increasing $w$ improves alignment but reduces MM. At $w\!=\!6$, our model reaches 32.2\% R@1 versus 16.6\% for retrieval, retains 0.589 MM, and approximately matches corpus-level diversity.

\section{Discussion and Future Work}
\label{sec:discussion}
Most ArtEmis descriptions express affect without specifying pose, lighting or
framing, limiting direct supervision for each component. The corpus is
dominated by portraits and therefore covers a narrow range of poses. Field of
view is assumed rather than estimated, multiple figures can be placed closer
together than in the source paintings, and the recovered lighting represents
only an approximate low frequency cue. 

Future work could weight training by
lighting confidence and use film data to broaden pose and shot distributions.
We also plan a study with professional illustrators comparing our stagings
with pose libraries and image search.

\bibliographystyle{ACM-Reference-Format}

\end{document}